\documentclass[10pt,twocolumn,letterpaper]{article}

\usepackage{cvpr}
\usepackage{times}
\usepackage{epsfig}
\usepackage{graphicx}
\usepackage{amsmath}
\usepackage{amssymb}
\usepackage{xcolor}
\usepackage[export]{adjustbox}

\usepackage{stfloats}

\usepackage[pagebackref=true,breaklinks=true,letterpaper=true,colorlinks,bookmarks=false]{hyperref}

\cvprfinalcopy 

\def\cvprPaperID{****} 

\ifcvprfinal\fi

\definecolor{correct}{rgb}{0.0,0.5,0.0}
\definecolor{wrong}{rgb}{0.8,0.0,0.0}
\definecolor{todo}{rgb}{7,.0,.3} 

\begin{document}









\title{Copy and Paste: A Simple But Effective Initialization Method for Black-Box Adversarial Attacks}









\author{
\begin{tabular}[t]{c}
Thomas Brunner\textsuperscript{1,2}
\and
Frederik Diehl\textsuperscript{1,2}
\and
Alois Knoll\textsuperscript{2}
\end{tabular} \\
\begin{tabular}[t]{c}
\textsuperscript{1} fortiss GmbH
\and
\textsuperscript{2} Technical University of Munich
\end{tabular}\\
\begin{tabular}[t]{c}
{\tt\small \{brunner,diehl\}@fortiss.org}
\and
{\tt\small knoll@in.tum.de}
\end{tabular}
}

\maketitle
\thispagestyle{empty}

\begin{abstract}
	Many optimization methods for generating black-box adversarial examples have been proposed, but the aspect of initializing said optimizers has not been considered in much detail. We show that the choice of starting points is indeed crucial, and that the performance of state-of-the-art attacks depends on it. First, we discuss desirable properties of starting points for attacking image classifiers, and how they can be chosen to increase query efficiency. Notably, we find that simply copying small patches from other images is a valid strategy. We then present an evaluation on ImageNet that clearly demonstrates the effectiveness of this method: Our initialization scheme reduces the number of queries required for a state-of-the-art Boundary Attack by 81\%, significantly outperforming previous results reported for targeted black-box adversarial examples. 

\end{abstract}


\section{Introduction}

\begin{figure}[htbp]
	
	\centering
	
	\begin{tabular}{cc}
		
		
		\vspace{-0.02in}
		
		\footnotesize{Original: \textbf{\textcolor{correct}{Ford Model T}}}
		&
		\footnotesize{Original: \textbf{\textcolor{correct}{Wombat}}}
		\\
				\vspace{0.01in}		
		\includegraphics[width=.21\textwidth]{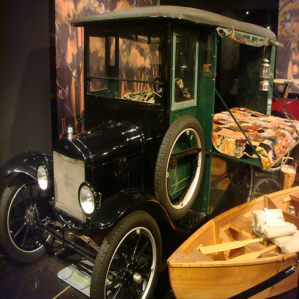}
		&
		\includegraphics[width=.21\textwidth]{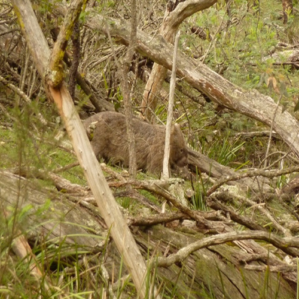} \\
		

		\vspace{-0.02in}
		\footnotesize{Starting point: \textbf{\textcolor{wrong}{Tibetan mastiff \ }}}&\footnotesize{Starting point:  \textbf{\textcolor{wrong}{Garter snake}}}\\
		
		\vspace{-0.06in}		
		\includegraphics[width=.21\textwidth]{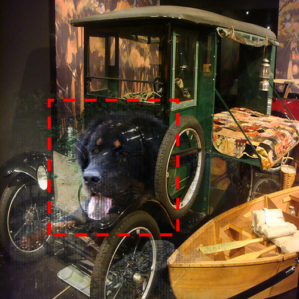}
		&
		\includegraphics[width=.21\textwidth]{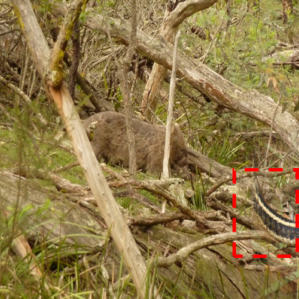} 
		\\

		\vspace{0.01in}		
		\footnotesize{$d_{\ell^2}= 36.22$} & 		\footnotesize{$d_{\ell^2}= 17.01$} \\


		\vspace{-0.02in}
		
		\footnotesize{Adv. example: \textbf{\textcolor{wrong}{Tibetan mastiff \ }}}&\footnotesize{Adv. example:  \textbf{\textcolor{wrong}{Garter snake}}}\\
		
		\vspace{-0.06in}
		\includegraphics[width=.21\textwidth]{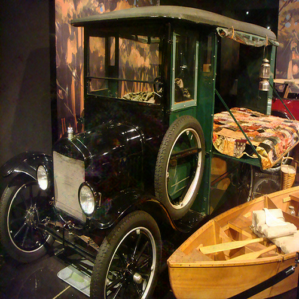}
		&
		\includegraphics[width=.21\textwidth]{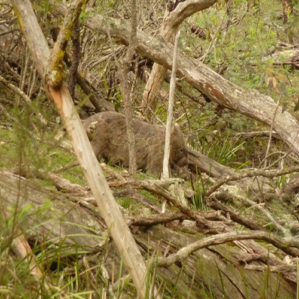} 
		\\
		
		\vspace{0.02in}	
		\footnotesize{$d_{\ell^2}= 9.04$} & 		\footnotesize{$d_{\ell^2}= 1.47$} \\

	\end{tabular}
	
	\caption{Starting points for adversarial attacks, synthesized by our method. (Top) Original images. (Middle) Starting points obtained by inserting small patches from images of the target class. (Bottom) Adversarial examples obtained after refining the images with a targeted black-box attack. Less than 1000 model queries are needed to render the perturbation virtually imperceptible, while the adversarial label is retained.}
	\label{fig:dog}	
	\vskip -0.125in
\end{figure}

Black-box adversarial attacks describe a scenario in which an attacker has access only to the input and output of a machine learning model, but no specific knowledge about its parameters or architecture \cite{Papernot:2017:PBA:3052973.3053009}. Intuitively, it may seem improbable for an attacker to simply guess the vulnerabilities of a model, but recent work has demonstrated that it is indeed possible to create custom-tailored adversarial examples for any black box when allowed to query the model a large number of times \cite{brendel17,ilyas18a,Ilyas2018PriorCB,autozoom,zoo}.

Naturally, the practicality of such attacks depends on the number of queries required and much work has gone into designing efficient optimization strategies.  Gradient estimation techniques have been popular \cite{ilyas18a, zoo, autozoom} for models that provide real-valued output (\eg softmax activations), and sophisticated sampling strategies have been proposed that drastically reduce the number of iterations \cite{Ilyas2018PriorCB}. The same has happened in the much harder label-only setting, where models only output a single discrete value (\eg the top-1 class label). Early success in this setting was sparked by the Boundary Attack \cite{brendel17}, which essentially performs a random walk along the decision boundary, and several variants have been proposed that improve the efficiency of this search procedure \cite{guo2018low,brunner18guessing}. Another family of successful black-box attacks exploits the fact that machine learning models often share vulnerabilities. They train surrogate models, perform white-box attacks and find adversarial examples that can be transferred to the model under attack \cite{madry_towards_2017,tramer17ens,Papernot:2017:PBA:3052973.3053009}. 

It is evident that considerable effort has gone into the design of sophisticated optimization procedures. Yet surprisingly, the question of how to initialize them has not been discussed in detail. We consider this to be a rather important gap in current literature, since costly optimization procedures can often be sped up by a smart choice of starting points. Our contribution is as follows:
\begin{itemize}
	\item We discuss beneficial properties of starting points and how they can improve the efficiency of iterative black-box attacks.
	\item As a proof of concept, we propose a simple copy-and-paste scheme, in which patches from images of the adversarial class are added to the image under attack.
	\item We use this strategy to initialize a state-of-the-art attack and evaluate it against an ImageNet classifier. Our initialization reduces the number of queries by 81\% when compared with previous results, and thus forms a new state of the art in query-efficient black-box attacks. The source code for repeating our experiment is publicly available \footnote{  \url{https://github.com/ttbrunner/blackbox_starting_points} }.
\end{itemize}

In this work, we exclusively focus on the targeted label-only black-box setting, where an attacker must change the classification to a specific label and does not have access to gradients or confidence scores. This is one of the hardest settings currently considered \cite{brendel17,brunner18guessing,chengHardlabel,ilyas18a} and therefore our results should be valid for easier settings as well. 


\section{Initialization strategies}

Currently, black-box adversarial attacks start with either (a) the original image or (b) an example of the target class. In the case of (a), the attack tries to take steps into directions that lead to an adversarial region. This is considered very hard and is typically approached by estimating gradients \cite{ilyas18a, zoo} or transferring them from a surrogate \cite{madry_towards_2017}. This approach can be unreliable and many queries must be spent before the adversarial region is found \cite{brunner18guessing}. In order to improve reliability, other attacks \cite{brendel17, ilyas18a, brunner18guessing} employ (b), where the starting point already has the desired class label but the distance to the original image is  high. As a result, the attack needs to travel a great distance through the input space, requiring many steps until it arrives at an example that is reasonably close to the original.

Both strategies have potential for improvement. For (a), Tram\`{e}r et al. \cite{tramer17ens} propose adding small random perturbations to the input, which they find to increase the overall success rate. In this work, we focus on (b) --  recent black-box attacks have achieved impressive results using this method \cite{brunner18guessing,ilyas18a}, and at the same time it seems very easy to improve. Surely some images of the target class would be better suited than others, and the number of required queries could be reduced by choosing them in a systematic manner.

\subsection{Criteria for suitable starting points}

In image classification, an adversarial example is considered successful if has low distance to the original image (\eg measured by some norm of the perturbation) and is at the same time classified as an adversarial label. We assume two properties to be beneficial for attack efficiency:

\textbf{Starting points should be close}. Intuitively, it makes sense to pick points that are already close to that goal. The optimization procedure would then merely refine them, requiring less iterations to arrive at an adversarial example or produce a better one in the same number of steps. The most straightforward approach is to search a large data set (\eg ImageNet) for images of the target class and then pick the one with the lowest distance to the original.

\textbf{Starting points should reduce dimensionality}. Optimization often suffers from very large search spaces. An ImageNet example at a resolution of 299 x 299 has 268,203 dimensions. It is therefore desirable to reduce this search space and to concentrate only on specific dimensions. Notably, Brunner et al. \cite{brunner18guessing} demonstrate that attacks gain efficiency by concentrating only on pixels that differ between the current image and the original, ignoring the rest. A suitable starting point should facilitate this, ideally by not replacing the original entirely but only small regions of it. The attack can then be limited to these regions.



\section{Copy-pasting adversarial features}

In order to test our assumptions, we propose a simple strategy that segments images of the target class into small patches and then inserts them into the image under attack. It is geared towards attacks targeting the $\ell^2$ norm, but similar strategies could be applied to improve $\ell^\infty$ attacks. This work should be understood as a proof of concept that offers many opportunities for refinement. Nevertheless, our evaluation in Section \ref{sec_experiments} shows that this simple approach already delivers a large boost in efficiency.

\subsection{Segmentation by saliency}

\begin{figure}[bp]
	
	\centering
	\footnotesize{(a)} \includegraphics[width=.13\textwidth]{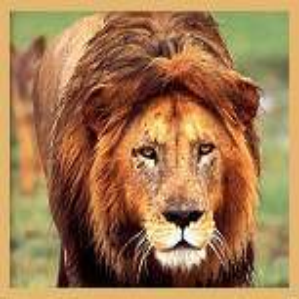}
	\hfill
	\footnotesize{(b)} \includegraphics[width=.13\textwidth]{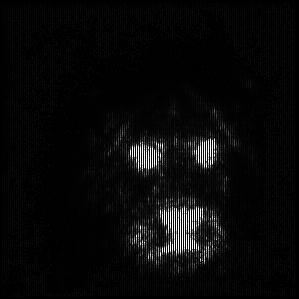}
	\hfill
	\footnotesize{(c)} \includegraphics[width=.13\textwidth]{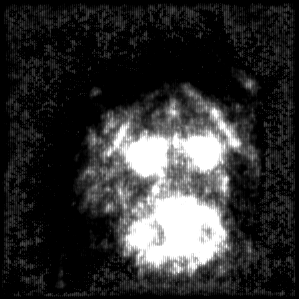}	
	\ \\
	\ \\
	
	\footnotesize{(d)} \includegraphics[width=.13\textwidth]{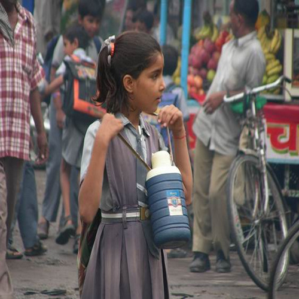}
	\hfill
	\footnotesize{(e)} \includegraphics[width=.13\textwidth]{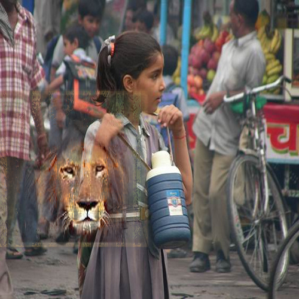}
	\hfill
	\footnotesize{(f)} \includegraphics[width=.13\textwidth]{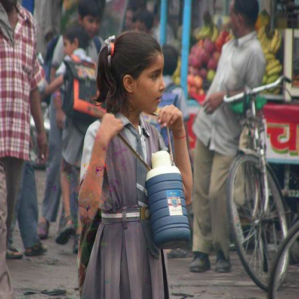}
	
	\caption{Segmentation of starting points. (a) An image of the target class, label "lion". (b) Saliency mask on a surrogate. (c) Smoothened and amplified mask to improve reliability. (d) Original image under attack, label "water bottle". (e) Starting point after the patch is inserted. (f) Final adversarial example.}
	\label{fig:blending2}	
\end{figure}

\begin{figure*}[t]

	\begin{tabular}{p{.14\textwidth}p{.14\textwidth}p{.14\textwidth}p{.14\textwidth}p{.14\textwidth}p{.14\textwidth}}
		\centering$i=1$
		&
		\centering$i=500$
		&
		\centering$i=1000$
		&
		\centering$i=2500$
		&
		\centering$i=5000$
		&
		\ \ \ \ \ $i=10000$\\
		
		\includegraphics[width=.15\textwidth]{img/girl-start-full.png}
		
		&
		\includegraphics[width=.15\textwidth]{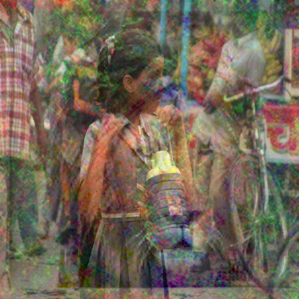}
		&	
		\includegraphics[width=.15\textwidth]{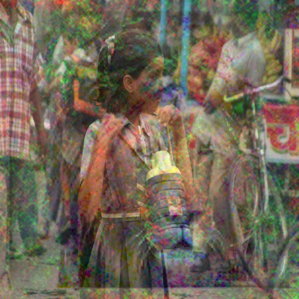}
		&		
		\includegraphics[width=.15\textwidth]{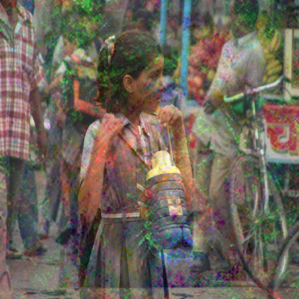}
		&			
		\includegraphics[width=.15\textwidth]{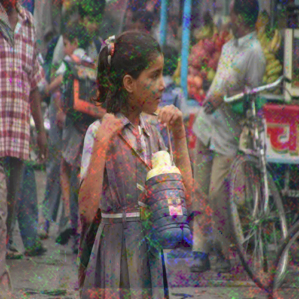}
		&			
		\includegraphics[width=.15\textwidth]{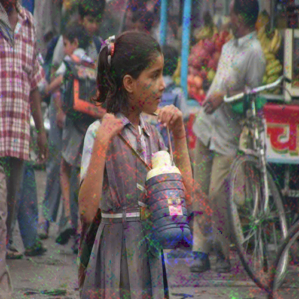}
		
		\\
		
		\vspace{-0.31in}	\centering$d_{\ell^2}=152.68$
		&
		\vspace{-0.31in}	\centering$d_{\ell^2}=55.30$
		&
		\vspace{-0.31in}	\centering$d_{\ell^2}=52.86$
		&
		\vspace{-0.31in}	\centering$d_{\ell^2}=41.38$
		&
		\vspace{-0.31in}	\centering$d_{\ell^2}=29.27$
		&
		\vspace{-0.31in}	\ \ \ \ \ $d_{\ell^2}=22.37$\\
		
		\vspace{-0.25in}	\includegraphics[width=.15\textwidth]{img/girl-start.png}
		&
		\vspace{-0.25in}	\includegraphics[width=.15\textwidth]{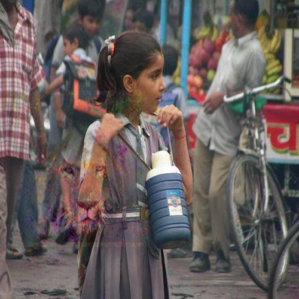}
		&
		\vspace{-0.25in}	\includegraphics[width=.15\textwidth]{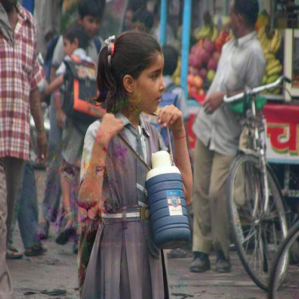}
		&	
		\vspace{-0.25in}	\includegraphics[width=.15\textwidth]{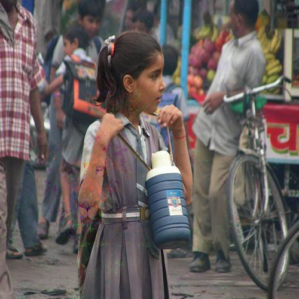}
		&		
		\vspace{-0.25in}	\includegraphics[width=.15\textwidth]{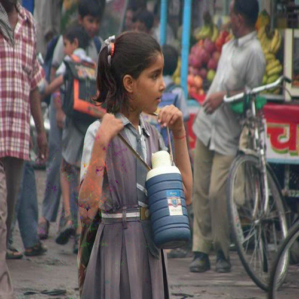}
		&			
		\vspace{-0.25in}	\includegraphics[width=.15\textwidth]{img/girl-start-10000.png}\\
		
		\vspace{-0.14in}	\centering$d_{\ell^2}=29.51$
		&
		\vspace{-0.14in}	\centering$d_{\ell^2}=14.84$
		&
		\vspace{-0.14in}	\centering$d_{\ell^2}=11.62$
		&
		\vspace{-0.14in}	\centering$d_{\ell^2}=7.93$
		&
		\vspace{-0.14in}	\centering$d_{\ell^2}=6.19$
		&
		\vspace{-0.14in}	\ \ \ \ \ $d_{\ell^2}=5.50$\\
		
	\end{tabular}
	
	\caption{Number of queries vs. perturbation magnitude for a Boundary Attack. (Top) Initialized with a full-sized image of the target class. (Bottom) Initialized with our method. All images are classified as the adversarial label, "lion".}
	\label{fig:comp}	
	\vskip -0.1in
\end{figure*}

The most important pixels for classification are those that contain salient features. They typically concentrate in small regions (\eg nose and eyes of an animal, see Figure \ref{fig:blending2}), whereas the rest of an image matters little for the predicted class label. The unimportant regions can safely be removed -- compare the starting points in Figure \ref{fig:comp} (left), where removing the background significantly lowers the $\ell^2$-distance to the original but retains the adversarial label.
To do this, we can apply any segmentation method of our choice. In our implementation, we use a surrogate model to construct saliency maps for images of the target class and then blend these pixels into the original image. 

 Saliency maps are model-specific, and therefore our initialization could be interpreted as a transfer attack that is not guaranteed to generalize across models. To address this concern, we apply heavy smoothing and amplification to the map. This results in contiguous patches that cover the salient regions and are therefore likely to contain the core motif of an image (see Figure \ref{fig:blending2}). We expect this method to generalize well across models, but in practice it can also be replaced by any other segmentation technique available to the attacker.

\begin{table*}[t]
	\begin{center}
		\begin{small}
			\begin{sc}
				\begin{tabular}{r|l|l|cccccc|c}
					\multicolumn{3}{l|}{} & \multicolumn{6}{c|}{Success rate vs number of queries} & Median queries \\
					\# & Attack & Initialization & 500 & 1000 & 2500 & 5000 & 10000 & 15000 & until success \\
					\hline
					1 & BBA \cite{brunner18guessing} & Closest image & 0.04 & 0.09 & 0.25 & 0.46	& 0.72 & 0.80 & 5485 \\
					
					2 & BBA \cite{brunner18guessing} & Copy and paste (ours) & 0.29 & 0.38 & 0.63 & 0.75 & 0.88 & 0.90 & 1541 	\\					
					
					
					3 & BBA (tuned) & \textbf{Copy and paste (ours)} & \textbf{0.32} & \textbf{0.49} & \textbf{0.74} & \textbf{0.86} & \textbf{0.94} & \textbf{0.96} & \textbf{1028} 	\\					
					
					
				\end{tabular}
			\end{sc}
		\end{small}
	\end{center}
	\vskip -0.00in
	\caption{Comparison of initialization strategies for a targeted label-only Boundary Attack on ImageNet. All numbers include queries made during initialization. Run 1 is the biased Boundary Attack (BBA) as implemented by Brunner et al. \cite{brunner18guessing}. Run 2 replaces their starting points with ours, but otherwise performs the same attack. In run 3, we tune some of the attack hyperparameters to take better advantage of our starting points, resulting in an even larger performance gain.}	
	\label{table:results-ba}
\end{table*}

\subsection{Placing adversarial patches}

It is apparent that the inserted pixels constitute small adversarial patches that change the classification of an entire image.  This effect has been described by Brown et al. \cite{Brown2018AdversarialP}, who construct small patches that are very salient but also strikingly visible to human observers. It would be possible to use their patches as starting points, however their strong visibility is contrary to our goals -- attacks in our setting aim to reduce the perturbation as much as possible in an attempt make the patches completely imperceptible to humans.

We use a simple brute-force method for placement. Salient patches are extracted from multiple images of the target class, randomly scaled and translated, and then blended over the original image. We create 50 such candidates and rank them by distance to the original image. The candidates are then tested against the black box and the first image to be adversarially classified is chosen as the final starting point. 

We typically find success within the first 10 candidates, but in the case that all are unsuccessful on the black-box model, an attacker may wish to fine-tune the procedure, increase patch size or fall back to full-sized images. In our evaluation, we are able to synthesize valid starting points for all examples.

\section{Evaluation}
\label{sec_experiments}

 To demonstrate the speedup provided by our initialization method, we apply it to a black-box attack against a pre-trained Inception-v3 ImageNet classifier \cite{inceptionv3}. 
 
 \textbf{Setting}. We choose a query-limited targeted label-only setting, which is currently considered to be one of the most difficult scenarios \cite{brendel17} to attack. We do not consider untargeted attacks, as ImageNet contains 1000 classes: An attacker could simply substitute one dog breed for another and thus create an "adversarial" example.

\textbf{Data set}. We randomly pick 1000 images from the ImageNet validation set and resize them to 299 x 299 x 3. For each image, we pick a random class label as the adversarial target. Our initialization method chooses images of the target class from the ImageNet validation set and uses a pre-trained Inception-ResNet \cite{Szegedy2016Inceptionv4IA} to extract salient regions. 
 
 \textbf{Attack method}. We perform a Boundary Attack \cite{brendel17}, which is one of the simplest $\ell^2$ attacks and at the same time has recently been shown to be one of the most query-efficient \cite{avc2018,brunner18guessing} in our setting. We use the publicly available implementation of Brunner et al. \cite{brunner18guessing}, who perform biased sampling to obtain state-of-the-art performance. We replace their initialization (closest image of the target class) with our own and activate their low-frequency and regional masking biases. Low-frequency perturbations have been shown to improve efficiency \cite{guo2018low,Ilyas2018PriorCB}, and the masking bias should be able to take full advantage of a low-dimensional search space. For the sake of simplicity we deactivate the surrogate bias, and use our surrogate model only for extracting the (smoothened) saliency maps. 
 
 \textbf{Success criteria}. We measure attack efficiency in number of queries until success. We define success as $d_{\ell^2} < 25.89$, which is the same threshold as used in the original attack \cite{brunner18guessing}. This number roughly corresponds to a worst-case $\ell^\infty$ perturbation of $0.05$ in our resolution, which is a threshold also used by other black-box attacks on ImageNet \cite{ilyas18a,Ilyas2018PriorCB}.

\textbf{Results}. Table \ref{table:results-ba} shows that our synthesized starting points greatly improve query efficiency, especially in the early stages of the attack.  Simply replacing the initialization method reduces the median number of queries from 5485 to 1541, which is a reduction by 72\%. Interestingly, some of the starting points synthesized by our method are already below the $\ell^2$ threshold (such as the snake in Figure \ref{fig:dog}, which was obtained in a single query). 

We perform another run with modified hyperparameters to take better advantage of our starting points (exact values are provided in our source code). This run achieves a median of 1028 queries, which in total amounts to a \textbf{reduction by 81\%} and, to the best of our knowledge, significantly outperforms the previous state of the art in targeted  black-box attacks. Figure \ref{fig:comp} shows a side-by-side comparison of attack progress for a single example.

\section{Conclusion}

We have shown that the performance of iterative black-box attacks greatly depends on their initialization. In a proof of concept, we have demonstrated how starting points can be synthesized by adding small patches to an image. Indeed, some of the images crafted by our method could be considered adversarial from the start, rendering the actual attack largely obsolete.

This seems to indicate that the limits of current benchmarks for evaluating black-box attacks are being reached, and that subsequent improvements might not add to their applicability in the real world. 
A better definition of "adversarial" is needed, and metrics such as $\ell^2$ (and, for that matter, $\ell^\infty$) should be replaced by measures based on human cognition. On high-resolution images, robustness should not be benchmarked by performing top-1 classification -- an image that visibly contains both a car and a dog (Figure \ref{fig:dog}) should not be considered adversarial. Future benchmarks could address this problem by including multi-object detection, or at least top-k classification.

Still, our discovery provides some pointers for future work on real-world  attacks. For example, an attacker may wish to create an adversarial patch as suggested by Brown et al. \cite{Brown2018AdversarialP}, and then use an iterative attack to camouflage it in the environment. It should also be interesting to see if adaptive attacks like the Boundary Attack can be modified to maintain a certain distance from the decision boundary, which would make the resulting examples robust in noisy scenarios and against non-deterministic classifiers. We consider this a prerequisite for finally bringing this family of attacks to the real world.
 
Our work shows that it is much easier to create black-box adversarial examples for high-resolution image classifiers than previously thought, and we hope this discovery provides a stepping stone towards a more realistic evaluation of black-box attacks in the future.
\clearpage

{\small
\bibliographystyle{ieee}
\bibliography{egbib}
}

\end{document}